\documentclass[10pt,conference,letterpaper]{IEEEtran}

\providecommand{\IEEESubmission}{1}

\usepackage[T1]{fontenc}
\usepackage[utf8]{inputenc}
\usepackage{amsmath}
\usepackage{amssymb}
\usepackage{graphicx}
\usepackage[font=footnotesize]{caption}
\usepackage{subcaption}
\usepackage{booktabs}
\usepackage{hyperref}
\usepackage[backend=biber,style=numeric-comp,sorting=none,maxbibnames=99]{biblatex}
\usepackage{color}
\usepackage{listings}
\usepackage{dblfloatfix}
\usepackage{etoolbox}

\graphicspath{{./}{figures/}{figures/rendered/}}

\newcommand{\swcdef}[1]{#1}
\newcommand{\swc}[1]{#1}

\begin{document}

% ============================================================
% Title block (English)
%   Abstract / Keywords text is defined once and reused in both
%   IEEEtran (\maketitle) and the legacy banner layout.
% ============================================================

\newcommand{\PaperTitleEn}{An Exposure-Time-Aligned Primary-Path Architecture for Autonomous-Driving ECUs}
\newcommand{\AbstractTextEn}{%
While end-to-end (E2E) autonomous driving has become the dominant research direction, production vehicles continue to rely on modular multi-NN pipelines for a non-trivial transitional period. The subject of this paper is the design of an architecture that, during this phase, supports a modular pipeline and an E2E path side by side and embeds a path for staged migration. Transplanted to a production SoC, egalitarian late fusion is compute-inefficient and offers no natural unit for staged E2E substitution. As an alternative, we propose three design principles: \textbf{(i) Primary-Path}, which selects a primary perception-to-planner chain and prioritizes its enclosure within a single SoC over the non-critical paths, \textbf{(ii) Exposure-Time-Aligned}, which propagates the primary sensor's exposure time $\tau_{\text{exp}}$ as a tag along the chain and event-drives the fusion node on matched $\tau_{\text{exp}}$ rather than a fixed cycle, and \textbf{(iii) Co-Path Coexistence}, which, building on (i) and (ii), lets an E2E output path co-run with the modular pipeline within the same $\tau_{\text{exp}}$ cycle. On a Dual-SoC production AD-ECU, the implementation achieves a camera-shutter to planner-output latency of 296 ms on average within a 350 ms design budget, with the modular pipeline and the E2E path coexisting within the shared $\tau_{\text{exp}}$ cycle.%
}
\newcommand{\KeywordsEn}{Autonomous driving, primary-path design, exposure-time alignment, modular-to-E2E transition, production AD-ECU}

\ifdefined\IEEESubmission
  % IEEEtran title block
  \title{\PaperTitleEn}
  \ifdefined\BlindSubmission
    \author{\IEEEauthorblockN{Anonymous Authors}\IEEEauthorblockA{Anonymous Affiliations}}
  \else
    \author{%
      \IEEEauthorblockN{Toru Saito\IEEEauthorrefmark{1}, Yuki Hagura\IEEEauthorrefmark{1}, Tatsuya Konishi\IEEEauthorrefmark{2}, Satoru Mizusawa\IEEEauthorrefmark{1}, Takumi Yajima\IEEEauthorrefmark{1}}%
      \IEEEauthorblockA{\IEEEauthorrefmark{1}Sony Honda Mobility Inc.\quad\IEEEauthorrefmark{2}Honda Motor Co., Ltd.}%
    }
  \fi
  \maketitle
  \begin{abstract}
  \AbstractTextEn
  \end{abstract}
\else
  % Legacy banner layout
  \begin{center}
  {\Large\bfseries \PaperTitleEn}

  \vspace{1cm}

  \ifdefined\BlindSubmission
    Anonymous Authors

    \vspace{0.3cm}

    Anonymous Affiliations
  \else
    Toru Saito\textsuperscript{1}, Yuki Hagura\textsuperscript{1}, Tatsuya Konishi\textsuperscript{2}, Satoru Mizusawa\textsuperscript{1}, Takumi Yajima\textsuperscript{1}

    \vspace{0.3cm}

    \textsuperscript{1}Sony Honda Mobility Inc. \quad \textsuperscript{2}Honda Motor Co., Ltd.
  \fi
  \end{center}

  \vspace{2mm}
  \noindent\textbf{Abstract}\hspace{0.5em}\AbstractTextEn
  \par\vspace{1mm}

  \vspace{3mm}\noindent\rule{\linewidth}{0.4pt}\vspace{1mm}
\fi

\section{Introduction}

Single-model end-to-end (E2E) autonomous driving has become the leading research direction (UniAD~\cite{hu2023uniad}, VAD~\cite{jiang2023vad}, GenAD~\cite{zheng2024genad}, DriveLM~\cite{sima2024drivelm}). Yet production vehicles retain modular multi-NN pipelines during the transition for practical reasons: (a) partitioned multi-team / multi-supplier development that does not match co-developing a monolithic model; (b) regulation/feature/sensor additions delivered most naturally as independent modules; (c) ISO 26262~\cite{iso26262} / ASIL safety processes assume functional decomposition with per-component verifiable requirements, which is more naturally satisfied by a modular pipeline than by a monolithic NN. The question is how to design the transitional modular architecture itself.

The standard starting point for modular perception fusion is late fusion in which peer perceptors are synchronized at a downstream fusion node, which in autonomous driving and robotics has become a de-facto practice through several open-source modular stacks~\cite{yurtsever2020adsurvey,fan2018apollo,kato2018autoware}. Its appeal is practical. Peer perceptors lack an obvious ordinal ranking and picking a primary is politically difficult under multi-supplier co-development with uncertain early-phase performance, and the symmetric layout keeps modules swappable. The initial architecture of the system studied here also used a \swcdef{EM-TimeSync} fusion node that waits, at a fixed cycle, for all perceptors to complete.

Egalitarian late fusion in this form interacts with how the production platform itself is partitioned. Production AD-ECU integration involves multiple compute domains separated by communication boundaries with non-deterministic latency (multi-SoC partitions, inter-ECU links over in-vehicle networks), and some segment of the deployment crosses such a boundary. We adopt a Dual-SoC AD-ECU as the target platform; on this configuration, a naive deployment of egalitarian late fusion was observed to incur three specific inefficiencies. The perception-to-planner chain gets split across the SoC boundary as peer perceptors saturate the sensor-I/O SoC, the fusion node is paced by the slowest branch, and peers offer no natural unit for staged E2E substitution. The empirical manifestation is unpacked in \S V.

We propose three interdependent design principles. \textbf{(i) Primary-Path}: a primary chain running from sensors through the dominant perceptor stack (here, the \swcdef{ViT Perception} stack, BEV fusion over Camera/LiDAR/Radar) to planner output is designated, where the perceptor selection aligns with the natural input/output boundary of a future single E2E model. The enclosure of this primary chain within a single SoC is prioritized over the non-critical paths formed by the remaining perceptors. \textbf{(ii) Exposure-Time-Aligned}: the primary sensor's exposure time $\tau_{\text{exp}}$ becomes the global reference, propagated as a tag along the chain, replacing the fixed-cycle wait at \swc{EM-TimeSync} with triggering on matched $\tau_{\text{exp}}$ and time-evolving asynchronous non-critical branches at the merge point. \textbf{(iii) Co-Path Coexistence}: on top of this base, a path from the BEV block inside \swc{ViT Perception} directly to \swcdef{NNPOST} (Fig.~\ref{fig:dataflow}) emits a trajectory alongside the modular pipeline under the same $\tau_{\text{exp}}$, so \swc{NNPOST} can compare and switch between them. The contribution of this paper is to quantify, on a Dual-SoC production AD-ECU, how egalitarian late fusion becomes inefficient when transplanted to such a platform and how the proposed principles restore deterministic latency while preserving an E2E migration path.

\section{Target System: SWC Layout and Data Flow of a Production AD-ECU}

\begin{figure}[t]
\centering
\includegraphics[width=\linewidth]{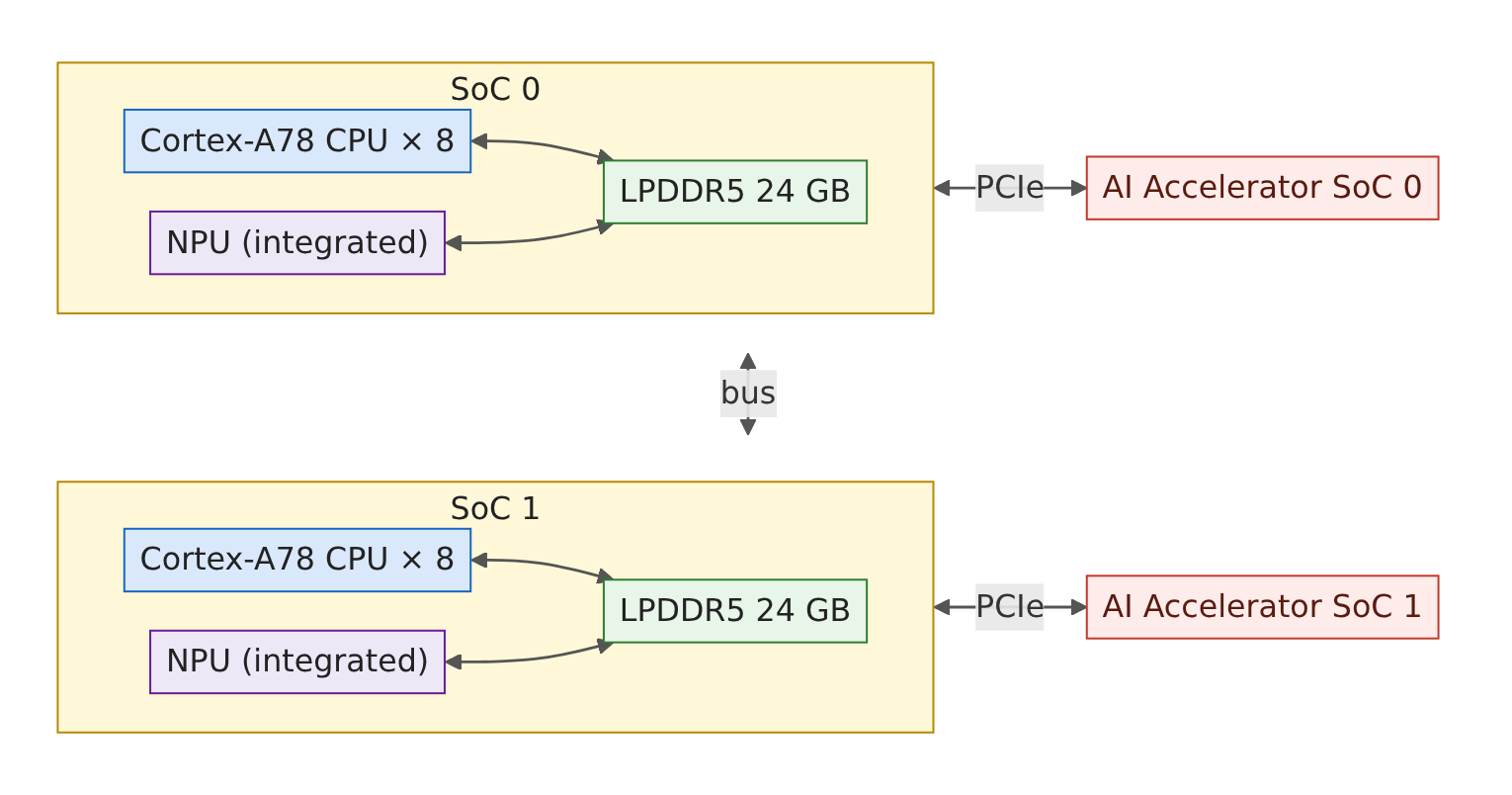}
\caption{Hardware layout of the target production AD-ECU (Dual-SoC pair). See §II for the per-SoC compute, memory, and interconnect details.}
\label{fig:hw}
\end{figure}

The target is a production AD-ECU (autonomous-driving electronic control unit) built around four SoCs: two general-purpose SoCs (SoC 0, SoC 1, hereafter the \textbf{Dual-SoC pair}), each integrating an 8-core Cortex-A78 CPU complex, an NPU, and 24 GB of CPU/NPU-shared LPDDR5; plus two inference-only AI-accelerator SoCs (no CPU) attached to the general-purpose SoCs over PCIe. SoC 0 and SoC 1 are linked by a general-purpose bus (Fig.~\ref{fig:hw}). The scope of this paper is the placement of CPU-driven software component (SWC) chains and the CPU/NPU pacing within and across the Dual-SoC pair; ``SoC 0 / SoC 1'', ``SoC-internal enclosure'', and ``cross-SoC'' refer to its boundary unless otherwise noted. Fig.~\ref{fig:dataflow} shows the resulting signal flow from sensor input to vehicle-control ECU.

\begin{figure*}[t]
\centering
\includegraphics[width=\linewidth]{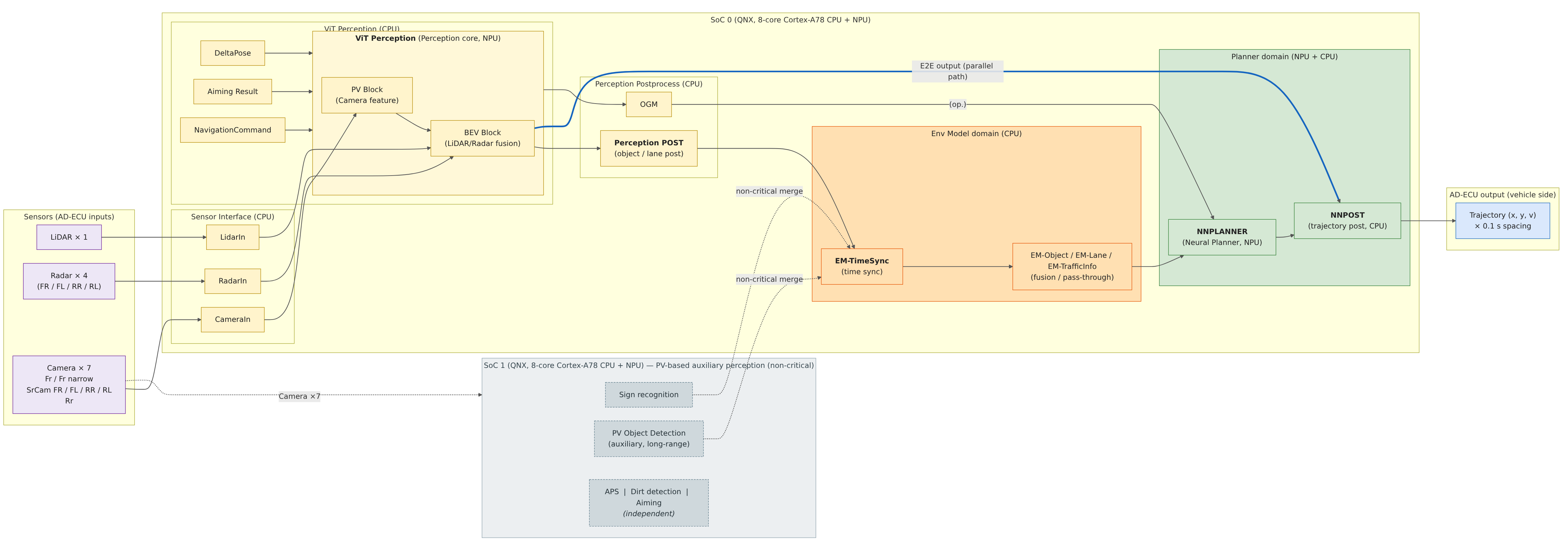}
\caption{Signal flow of the target production AD-ECU. Purple = sensors, yellow = perception domain (SoC 0, BEV-space), orange = env-model domain, green = planner domain, blue = vehicle-side ECU, grey = PV-based auxiliary perception (SoC 1, non-critical). CPU/NPU partitions inside each SoC are noted in the boxes; per-SoC compute details are in §II. Thin solid arrows form the \emph{modular critical path}; the bold blue edge is the parallel E2E path (BEV block inside \swc{ViT Perception} $\to$ \swc{NNPOST}, see §VII-A).}
\label{fig:dataflow}
\end{figure*}

\subsection{How to read Fig.~\ref{fig:dataflow}}

Every arrow between two SWCs is in fact middleware layers (SOME/IP, Zero-Copy DDS). The chain of SWCs and communication layers shown in the diagram is what this paper refers to as the critical path; the bold blue edge overlaying it is the parallel E2E path (\S I, \S VII-A).

The skeleton has three pieces. \textbf{Critical path}: the thin solid arrows in Fig.~\ref{fig:dataflow} trace the chain from sensors through \swc{ViT Perception}, the merge node \swc{EM-TimeSync}, the env-model SWCs, to \swc{NNPLANNER}/\swc{NNPOST}. This paper's proposals (SoC-internal enclosure, exposure-time alignment) target this chain and lock every SWC on it to the same 100 ms execution cycle on SoC 0 (per-cycle pacing; the end-to-end latency budget is separate and reported in §VI). The NPU runs the NN inference modules and the CPU hosts the post-processing, fusion, and orchestration SWCs, as marked in Fig.~\ref{fig:dataflow}. \textbf{Non-critical paths}: SoC 0 concentrates BEV-space perception, while SoC 1 hosts five parallel components in PV space: sign recognition, PV-based auxiliary object detection, automated parking system (APS), dirt detection, and aiming. Sign recognition and PV object detection merge asynchronously into the critical path at \swc{EM-TimeSync} via Ethernet/DDS, time-evolved to $\tau_{\text{exp}}$; the other three operate independently and do not connect to the merge point. \textbf{Control layer}: the motion controller consumes the \swc{NNPOST} trajectory (50 points at 0.1 s spacing over 5 s); this paper's scope ends at \swc{NNPOST} output. Inter-SWC communication is implemented as a combination of AUTOSAR Adaptive Platform ara::com and Zero-Copy DDS, with gPTP providing inter-ECU time synchronization. This forms the implementation foundation that underpins the deterministic merge of \S IV.

\section{Related Work and the Gap}

\textbf{E2E autonomous driving.} UniAD~\cite{hu2023uniad} (CVPR 2023 Best Paper) established query-based unified learning; VAD~\cite{jiang2023vad}, GenAD~\cite{zheng2024genad}, and DriveLM~\cite{sima2024drivelm} extend the line. DriveVLM~\cite{tian2024drivevlm} and Senna~\cite{jiang2024senna} propose configurations that run a VLM at low frequency alongside the E2E stack to use language reasoning as a supplementary signal. Evaluation in this line is mostly open-loop on nuScenes~\cite{caesar2020nuscenes} and does not cover concurrent execution, worst-case execution time (WCET), or scene-dependent load on production SoCs. Zhai \emph{et al.}~\cite{zhai2023openloop} reports a simple MLP matches several E2E methods on the same metric, questioning rankings built on this metric alone, though latency is not part of that critique. \textbf{Dual-track / safety-filter approaches.} A line of work constrains a learned planner with an independent rule-based or interpretable component: Lyft's SafetyNet~\cite{vitelli2021safetynet} runs an NN-planner output alongside a rule-based fallback layer, and Shao \emph{et al.}'s InterFuser~\cite{shao2022interfuser} uses interpretable intermediate features to constrain actions to a safe set, neither running two output paths synchronously in parallel.

\textbf{Gap.} Systematic surveys of end-to-end autonomous driving~\cite{chen2024e2esurvey,tampuu2020e2esurvey} focus on the algorithmic axis (learning methods, architectures, evaluation) and do not address the implementation methodology of multi-NN coexistence on production SoCs. On the other side, latency measurement work by Betz \emph{et al.}~\cite{betz2023latency1,betz2023latency2} focuses on empirical measurement methodology on ROS 2 stacks and does not enter the design-prescription side. A unified treatment of perception-fusion-planning placement, a global time reference, and an E2E migration path as a single design principle for production multi-NN environments falls outside the explicit scope of either line.

\section{Primary-Path / Exposure-Time-Aligned Design Principle}

\subsection{Terminology}

\textbf{Reference sensor}: the front camera; its exposure-window center is $\tau_{\text{exp}}$. \textbf{E2E latency budget $D_{\text{E2E}}$}: the allowable elapsed time from $\tau_{\text{exp}}$ to planner output ($\sim$350 ms here), set as a production target for the target ODD and allocated separately from the upstream sensor-capture and downstream motion-control budgets; deriving the budget value itself is out of scope. \textbf{Critical path}: the dominant NN/SWC chain from sensors to \swc{NNPOST} (\S II, Fig.~\ref{fig:dataflow}). \textbf{Non-critical paths}: NN/SWCs outside it whose outputs merge at \swc{EM-TimeSync}, time-aligned to $\tau_{\text{exp}}$ via coordinate transform and constant-velocity extrapolation.

\noindent\textbf{Architecture-level constraint.}\quad The critical path is trigger-driven: $\tau_{\text{exp}}$ frames are emitted at the regular cadence $T_{\text{exec}}$ (100 ms here), and each SWC is expected to complete within one cycle. The principles below are stated under this trigger-driven operation, and the upstream sensor-side prerequisites that keep the $\tau_{\text{exp}}$ rhythm regular are part of the architecture (\S IV-B).

We take the stance of deciding (a) which SoC encloses the critical path, then (b) where the merge point sits and to which time reference it aligns.

\subsection{The four principles}

\noindent\textbf{P1 Explicit critical-path definition.}\quad Before any re-placement, fix the node sequence $P^{*}$ that directly affects $\tau_{\text{out}}$. SoC placement and related decisions are evaluated against $P^{*}$.

\noindent\textbf{P2 Exposure time as the sole global reference promoted to the SWC layer.}\quad $\tau_{\text{exp}}$ is promoted from a sensor-side attribute to the timestamping and merge-condition reference visible across the SWC layer of the entire critical path. Concretely: (i) the $\tau_{\text{exp}}$ value of the originating sensor frame is carried as a tag on the data flowing through every SWC on the chain; (ii) each SWC is triggered on data arrival in the usual dataflow sense, but the frame is now uniquely identified by its $\tau_{\text{exp}}$ along the chain rather than by an SWC-internal time base; (iii) the merge point waits for inputs that share the same $\tau_{\text{exp}}$, replacing the conventional fixed-cycle wait. $\tau_{\text{exp}}$ is the only candidate that ties to the physical sampling moment and remains coherent across modular and E2E eras (an E2E NN can be designed to respond to the world state at $\tau_{\text{exp}}$).

\noindent\textit{Sensor-side timing-lock prerequisite.}\quad The per-cycle completion requirement of \S IV-A and the chain-wide $\tau_{\text{exp}}$ reference of P2 together require the sensor inputs to be physically locked to a common timing reference, in practice by triggering the cameras and the LiDAR so that their exposure and scan centers coincide. Conventional late fusion is comparatively robust to such upstream drift on the timing-design axis, since alignment is reconstructed at the fusion node from per-branch timestamps; the present architecture cannot rely on that margin. The sensor-side timing lock is therefore not an out-of-scope hardware detail but a prerequisite that follows from \S IV-A and P2. The mechanism by which a violation degrades the chain, and its empirical demonstration, are deferred to \S VI-C.

\noindent\textbf{P3 SoC-internal enclosure of the critical path.}\quad Inter-node communication on $P^{*}$ is routed over intra-SoC DDS (sub-ms, deterministic) so that non-deterministic cross-SoC routes are kept off the critical path, enabling deterministic WCET characterization. When platform constraints make a cross-SoC segment on $P^{*}$ unavoidable, the principle is to minimize and explicitly engineer the residual rather than to forbid the configuration.

\noindent\textbf{P4 Alignment by time evolution at merge points.}\quad Results arriving from non-critical branches (other SoCs, auxiliary perception, asynchronous sensors) are time-evolved (position and velocity) by the difference between their own timestamp and $\tau_{\text{exp}}$. This is implemented by fixing the fusion module (\swc{EM-TimeSync}) trigger to $\tau_{\text{exp}}$.

\section{Application to a Production Platform}

\subsection{Target platform}

The optimization target is the Dual-SoC pair of \S II. Each SoC runs QNX RTOS with AUTOSAR Adaptive, hosting more than 50 SWCs concurrently. Sensors comprise front/tele/rear/surround cameras, a LiDAR, five radars, and GNSS/IMU. The principal SWCs are listed in Table~\ref{tab:swc-trigger}.

\begin{table*}[t]
\centering
\caption{Principal SWCs on the target platform, with trigger differences between the Conventional Fusion and Proposed architectures.}
\label{tab:swc-trigger}
\footnotesize
\renewcommand{\arraystretch}{1.1}
\begin{tabular}{|p{3.5cm}|p{3.5cm}|p{4.5cm}|p{3.0cm}|}
\hline
Module & Role & Conventional Fusion & Proposed \\
\hline
\swc{ViT Perception} & Primary perception & data-arrival / $\tau_{\text{exp}}$ & data-arrival / $\tau_{\text{exp}}$ \\
\swc{Perception POST} & Object/Lane Tracking & data-arrival / $\tau_{\text{exp}}$ & data-arrival / $\tau_{\text{exp}}$ \\
\swc{OGM} & Occupancy grid & data-arrival / $\tau_{\text{exp}}$ & data-arrival / $\tau_{\text{exp}}$ \\
\swc{EM-TimeSync} & Exposure-time merge & Cyc 100 ms (fixed) / merge-time & data-arrival / $\tau_{\text{exp}}$ \\
\swc{EM-Object/Lane/TrafficInfo} & Fusion / Association & data-arrival / \swc{EM-TimeSync} time & data-arrival / $\tau_{\text{exp}}$ \\
\swc{NNPLANNER} & Neural planner & data-arrival / \swc{EM-TimeSync} time & data-arrival / $\tau_{\text{exp}}$ \\
E2E output & Parallel E2E path & Async (co-execution infeasible) & data-arrival / $\tau_{\text{exp}}$ \\
\swc{NNPOST} & Trajectory postprocess & data-arrival / \swc{EM-TimeSync} time & data-arrival / $\tau_{\text{exp}}$ \\
Auxiliary perception (PV) & Sign recognition / PV object detection & Async $\to$ time-aligned at \swc{EM-TimeSync} & Async $\to$ time-evolved to $\tau_{\text{exp}}$ \\
\hline
\end{tabular}
\end{table*}

\noindent\emph{Note.} Each cell shows ``trigger / frame timestamp''. The Conventional Fusion drives \swc{EM-TimeSync} on a fixed 100 ms cycle and stamps every downstream SWC with the merge time, losing $\tau_{\text{exp}}$ along the chain. The proposed configuration propagates $\tau_{\text{exp}}$ from \swc{ViT Perception} to \swc{NNPOST} and time-evolves async branches at the merge point.

\subsection{Conventional Fusion as a mapping of egalitarian late fusion}

The \textbf{Conventional Fusion} is the egalitarian late-fusion mapping we adopted in the early phase of the project (Fig.~\ref{fig:arch-compare}, top). The \swc{ViT Perception} chain and the PV auxiliary perception are co-located on SoC 0, \swc{EM-TimeSync} waits on a fixed cycle (Cyc), and $\tau_{\text{exp}}$ is carried only as a payload attribute. The resulting CPU saturation on SoC 0 forced the NNPlanner (\swc{NNPLANNER}/\swc{NNPOST}) onto SoC 1, so the perception-to-planner chain crosses the SoC boundary and is exposed to non-deterministic jitter on that cross-SoC segment.

\subsection{Inefficiencies exposed in the Conventional Fusion}

The starting point of egalitarian \emph{merge-by-waiting} surfaced two further inefficiencies on the critical path:

\begin{itemize}
\item[(1)] \textbf{Worst-case latency = normal + one period.} When \swc{EM-TimeSync} waits at a fixed cycle for every upstream arrival, a one-period delay of any branch propagates as a full one-period wait at the merge point, so the worst-case end-to-end latency is normal-case latency plus one full cycle. The slowest branch effectively dictates the budget, and faster branches are held back.

\item[(2)] \textbf{No natural E2E substitution unit.} With peer perceptors there is no obvious chunk to replace as a single E2E model later, so gradual migration lacks a consistent boundary.
\end{itemize}

\subsection{Proposed architecture with primary selection and $\tau_{\text{exp}}$ alignment}

\begin{figure*}[t]
\centering
\includegraphics[width=\linewidth]{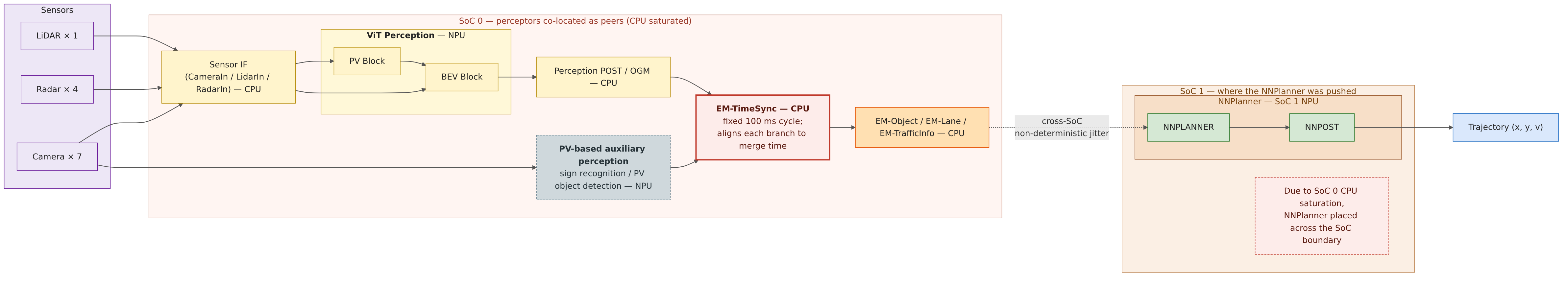}
\vspace{1mm}
\includegraphics[width=\linewidth]{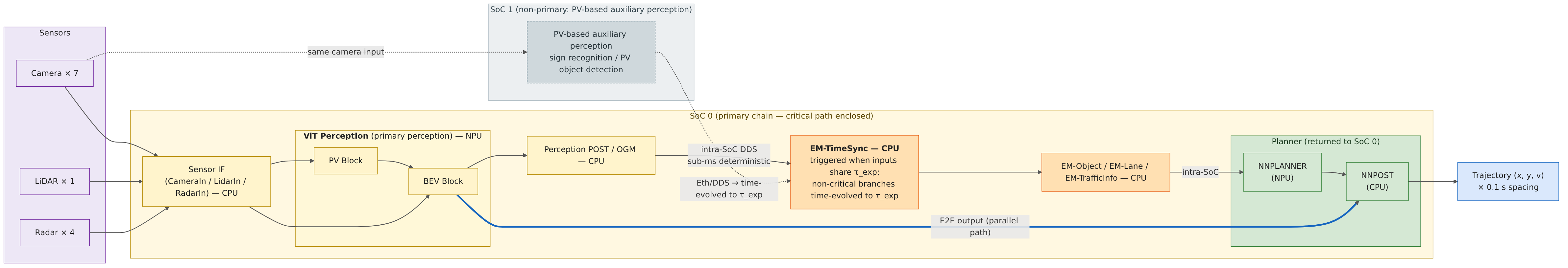}
\caption{Architecture comparison. \textbf{Top --- Conventional Fusion (Before)}: under egalitarian peer placement the \swc{ViT Perception} chain and the PV-based auxiliary perception are co-located on SoC 0; \swc{EM-TimeSync} waits at a fixed cycle (Cyc), the SoC 0 CPU is saturated, and the NNPlanner is pushed to SoC 1, so ``fusion $\to$ planning'' crosses the SoC boundary. \textbf{Bottom --- Proposed (After)}: the \swc{ViT Perception} chain is designated as the primary, and the entire critical path (perception $\to$ fusion $\to$ planning) is enclosed in SoC 0; \swc{EM-TimeSync} is pinned to the primary exposure time $\tau_{\text{exp}}$ and PV-based auxiliary perception remains on SoC 1 as a non-critical path that merges by time evolution.}
\label{fig:arch-compare}
\end{figure*}

The natural fix is to designate one such sensor-to-planner chain as the primary and position its end-to-end latency as the dominant factor (P1). We pick the chain running through the \swc{ViT Perception} stack because that stack produces the dominant subset of planner inputs (3D objects, occupancy, lanes) while auxiliary perceptors contribute add-on signals (signs, PV objects), and its BEV intermediate matches the natural I/O boundary of a future single E2E model (BEV $\to$ trajectory). The chain therefore becomes the absorption unit for staged migration, while non-critical perceptors remain independent.

Once the primary is fixed, the other decisions follow as corollaries (Fig.~\ref{fig:arch-compare}, bottom). The primary is enclosed in SoC 0 (P3); auxiliary perception sits on SoC 1 as a non-critical path. The reference time of the merge node is shifted from the merge-trigger time to the primary sensor's exposure time $\tau_{\text{exp}}$: the $\tau_{\text{exp}}$ tag is propagated through every SWC on the chain, and \swc{EM-TimeSync} replaces its fixed-cycle wait with a synchronized merge that pairs inputs sharing the same $\tau_{\text{exp}}$ (P2, P4). The NNPlanner returns to its natural position inside SoC 0 (CPU host, NPU compute), and SoC 0's CPU pressure is relieved by moving the auxiliary perception out.

On top of this, a parallel E2E path (Fig.~\ref{fig:dataflow}, bold blue) emits a trajectory directly from BEV features and runs alongside the modular chain in the same cycle, with \swc{NNPOST} comparing and selecting between them (a real-vehicle instance is shown in Fig.~\ref{fig:co-path}; details in \S VII-A).

This $\tau_{\text{exp}}$-keyed synchronized merge eliminates the fixed-cycle wait on the critical path (up to 100 ms) and confines kinematic extrapolation to the asynchronous non-critical branches; the critical chain reaches the merge point without any time-evolution step. The same $\tau_{\text{exp}}$ reference also persists across the later primary handover to a single E2E model, minimizing motion-controller adaptation.

\section{Validation: Production Implementation Results}

\subsection{Setup}

This section reports timing measurements of the Proposed package as implemented on the production AD-ECU and discusses how the observed behavior matches the design intent of P1--P4. Measurements capture $>$ 400 k events in Chrome Trace Event Format on the same Dual-SoC platform family as in \S II, allowing per-module start/stop times of every SWC on the critical path to be visualized. The system has a heterogeneous-period sensor setup (camera 33 ms, LiDAR 50 ms, execution period $T_{\text{exec}}$ = 100 ms). \swc{ViT Perception} decomposes into a per-camera perspective-view stage (\swc{PV Block}) that extracts image-plane features and a BEV-fusion stage (\swc{BEV Block}) that lifts the PV features into BEV space and fuses them with LiDAR/Radar; \swc{BEV Block} cannot start until the corresponding \swc{PV Block} outputs are available. The measurements below were taken with the sensor-side timing lock of \S IV-B in place; the counterpart behavior without it is shown in C.

\subsection{Critical-path timing under the Proposed package}

Fig.~\ref{fig:timelines} (top) shows a trace gantt of the proposed implementation over 500 ms. The Camera-shutter $\to$ \swc{NNPLANNER} chain repeats on a 100 ms cycle anchored at $\tau_{\text{exp}}$, with each SWC triggered on data arrival from the previous one. The entire chain closes within a single deterministic period. This is the critical path that P1--P4 prescribe, observed end-to-end on the production hardware.

\begin{figure}[t]
\centering
\includegraphics[width=\linewidth]{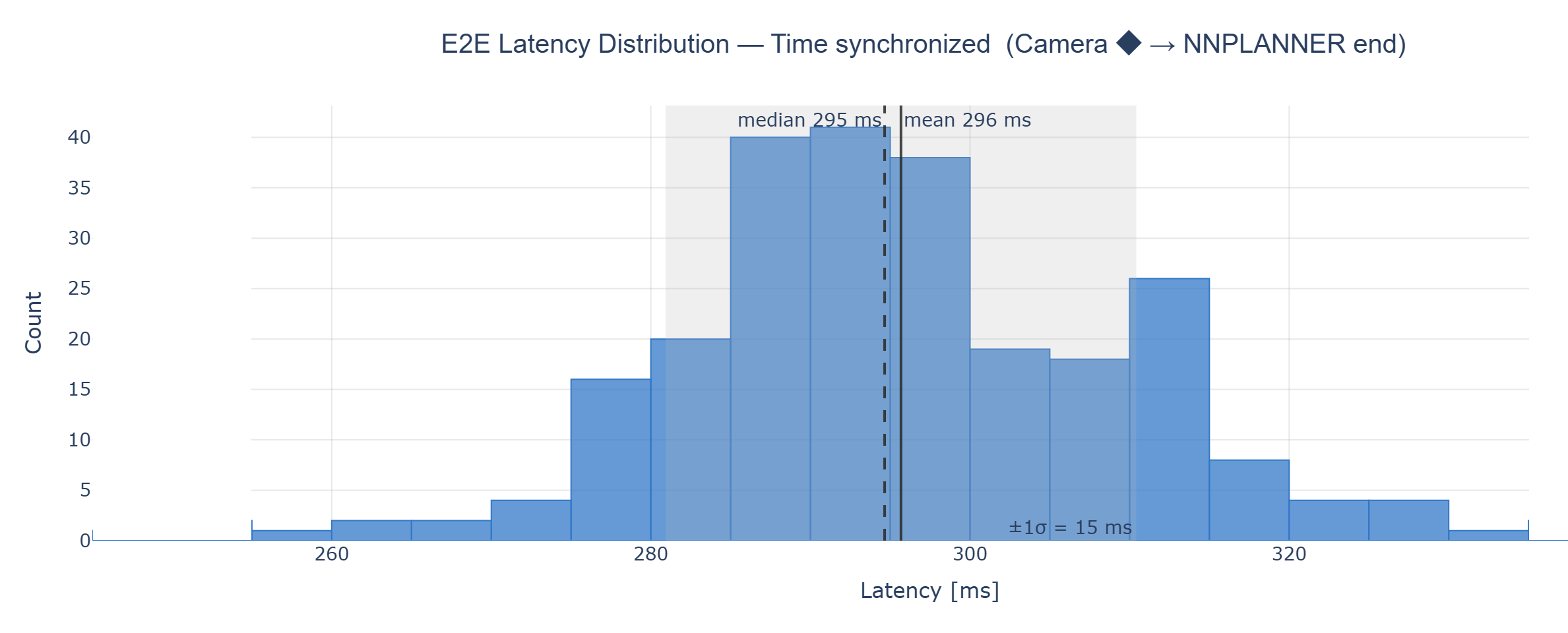}
\caption{Camera-shutter to \swc{NNPLANNER}-done latency distribution under the proposed package on the production AD-ECU. Median 295 ms, mean 296 ms, $\pm 1\sigma$ = 15 ms.}
\label{fig:latency-dist}
\end{figure}

The principal latency observation is the camera-shutter to \swc{NNPLANNER}-done latency distribution shown in Fig.~\ref{fig:latency-dist}, and the chain completes its synchronized merge within every 100 ms execution period. The merge node never waits for an out-of-cycle branch (worst case for fixed-cycle waiting fusion is normal + one period; that condition does not occur once $\tau_{\text{exp}}$-keyed synchronized merge and chain enclosure within SoC 0 are in place). The perception-to-planner chain no longer crosses the SoC boundary, removing the non-deterministic jitter that the cross-SoC segment would otherwise introduce.

\subsection{Empirical degeneration when the upstream timing lock is broken}

The deterministic timing of \S VI-B presupposes the sensor-side timing lock stated in \S IV-B. As noted there, conventional late fusion absorbs upstream drift at the fusion node and the present architecture cannot. P2 promotes $\tau_{\text{exp}}$ to a chain-wide reference, so any sensor-side drift accumulates into the per-cycle execution-time margin of every downstream SWC and eventually pushes some SWC out of its 1-cycle envelope.

\begin{figure*}[t]
\centering
\includegraphics[width=\linewidth]{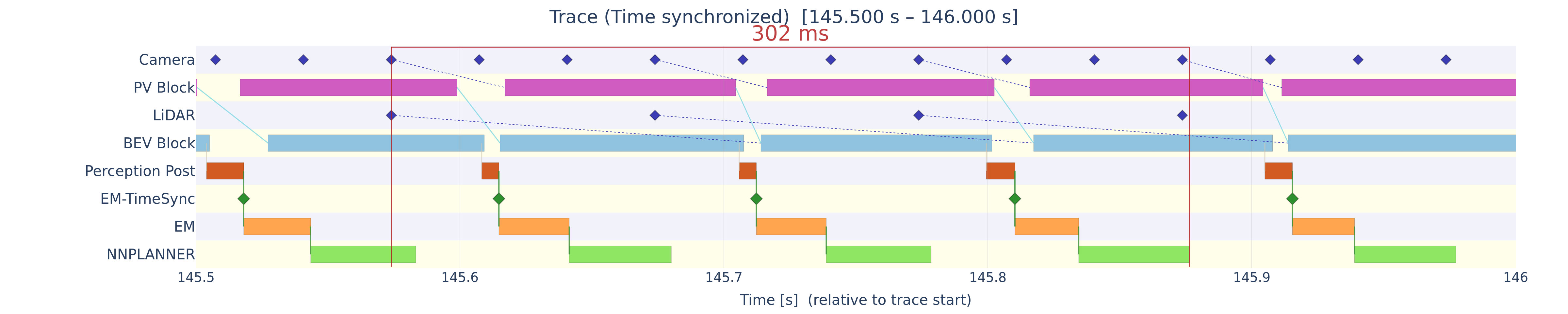}

\vspace{2mm}

\includegraphics[width=\linewidth]{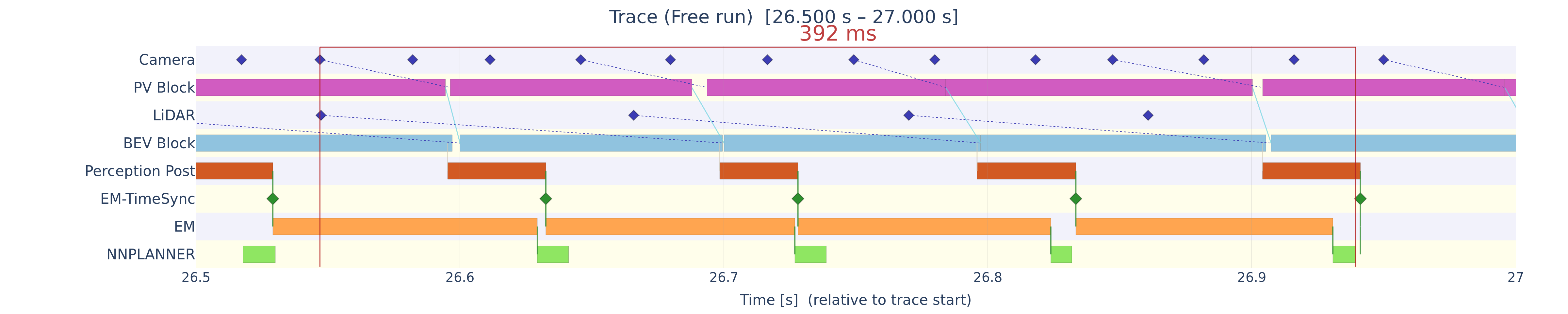}
\caption{Trace gantt over 500 ms on the production AD-ECU. \textbf{Top} (main result, \S VI-B): under the proposed package with the sensor-side timing lock of \S IV-B, Camera and LiDAR share $\tau_{\text{exp}}$, so \swc{PV Block} finishes by the time LiDAR arrives, \swc{BEV Block} starts without an idle gap, and \swc{Perception POST}, \swc{EM-TimeSync}, and \swc{NNPLANNER} chain through the rest of the 100 ms cycle. \textbf{Bottom} (\S VI-C): the same chain operated without the upstream timing lock, included as the principle-level counterpart. Each SWC is triggered with irregular phase offsets relative to the sensor inputs, pushing the \swc{BEV Block} start time later. The shift propagates downstream, and the chain loses its deterministic per-cycle behavior. The downstream SWCs no longer satisfy their individually allocated execution-time budgets and exhibit module-dependent failure modes, with prolonged execution in some and premature termination without normal completion in others.}
\label{fig:timelines}
\end{figure*}

Fig.~\ref{fig:timelines} (bottom) shows a representative trace of the chain operated without the upstream timing lock, included as the principle-level counterpart to the top panel.

\section{Discussion}

\subsection{Parallel modular / E2E operation for gradual migration}

\begin{figure}[h]
\centering
\includegraphics[width=\linewidth]{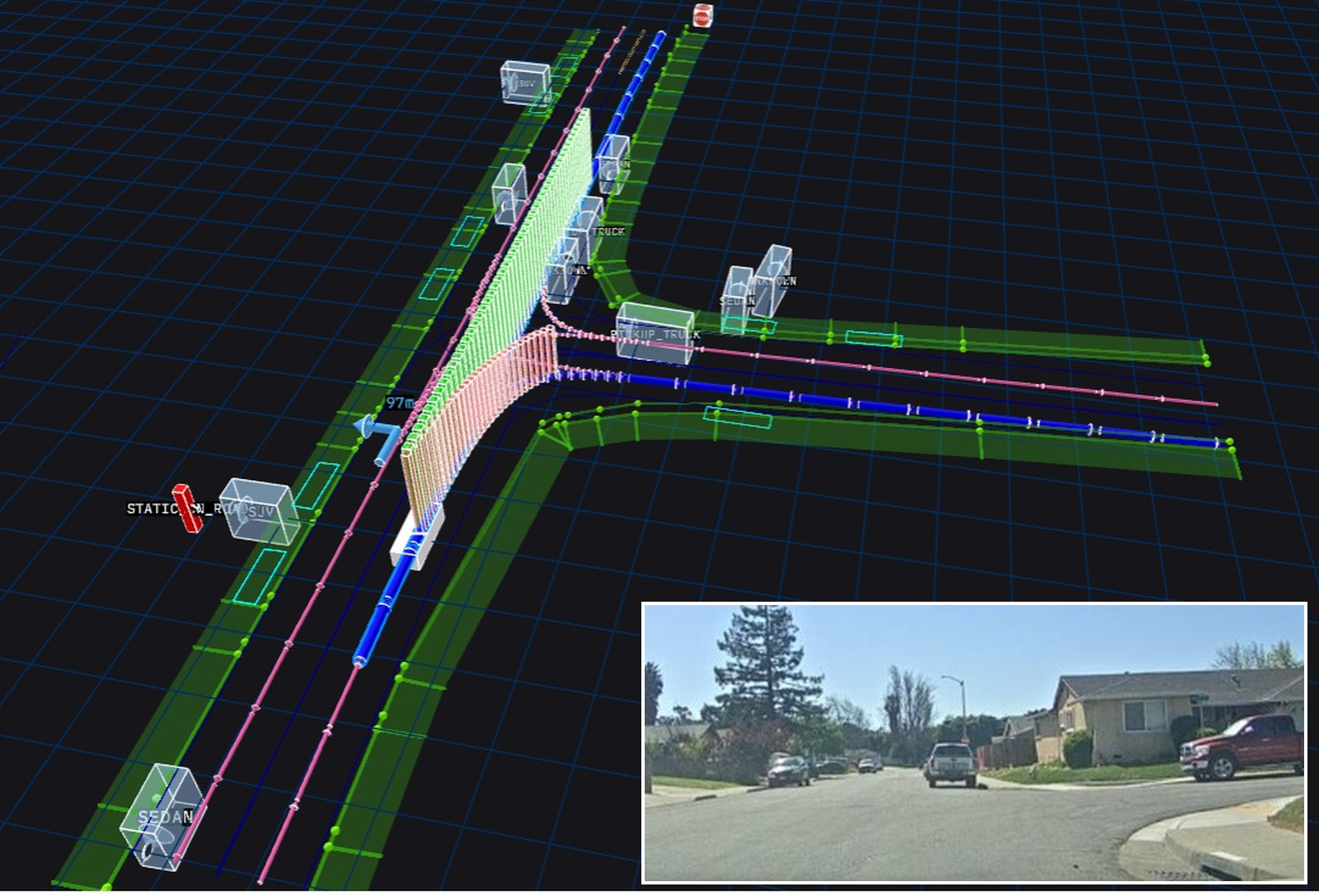}
\caption{Co-path coexistence on the production AD-ECU during real-vehicle operation. The 3D BEV view shows the modular \swc{NNPLANNER} trajectory (green) and the E2E trajectory (red) co-running within the same $\tau_{\text{exp}}$ cycle and arriving together at \swc{NNPOST}; the inset (lower right) is a reference camera view of the scene (not used as input to the AD-ECU). In this frame the green (modular) trajectory continues straight while the red (E2E) trajectory decelerates and turns right, so the two paths visibly diverge. This exemplifies the arbitration responsibility \swc{NNPOST} holds in \S V-D and this section.}
\label{fig:co-path}
\end{figure}

The proposed design lets the modular pipeline and an E2E path (BEV block inside \swc{ViT Perception} $\to$ \swc{NNPOST} direct) co-run within the same $\tau_{\text{exp}}$ cycle on top of the primary path (Fig.~\ref{fig:dataflow}). Fig.~\ref{fig:co-path} shows this co-run on the production AD-ECU during real-vehicle operation: the modular trajectory (green) continues straight while the E2E trajectory (red) decelerates and turns right, both reaching \swc{NNPOST} in the same cycle for arbitration. In the current deployment, the control output is taken from the E2E path during nominal operation, while the modular output is reserved for monitoring adherence to the navigation route and traffic rules; when the E2E trajectory diverges from these requirements, \swc{NNPOST} explicitly switches the lane selection to the modular branch.

As secondary uses, the same mechanism supports shadow evaluation of the E2E path against the modular primary, progressive expansion of E2E adoption, and an eventual handover of the primary role to E2E with the modular pipeline retained as a runtime monitor. Throughout, the $\tau_{\text{exp}}$ reference and the \swc{NNPOST} boundary are unchanged, so motion-controller adaptation across any such transition is minimal.

\subsection{Limitations and future work}

\begin{enumerate}
\item[(i)] \textbf{Generalization to other SoC layouts}: the principle is platform-neutral but the numeric outcome (mean 296 ms) depends on the CPU/NPU performance profile of the SoC family in use; re-derivation on different SoC layouts (other families, single-SoC integration, accelerator-side CPU configurations) is outside the scope of this paper.

\item[(ii)] \textbf{Controlled-setting timing analysis}: the measurements were taken under stationary and defined scenarios, and the cycle-by-cycle variation can grow further with scene-dependent load (e.g., perceptor post-processing cost with object and lane counts). Quantifying the scene-dependent variation under realistic traffic densities and bounding the worst case within the per-cycle execution period and budget is left to future work.

\item[(iii)] \textbf{No formal real-time proof}: a strict WCET upper bound for merge alignment is not given; connecting to formal real-time analysis is a natural next step.

\item[(iv)] \textbf{Parallel E2E evaluation}: \S VII-A is architectural; per-scene quantitative comparison is ongoing real-vehicle work, to be reported separately.
\end{enumerate}

\section{Conclusion}

We presented three design principles for the transitional modular architecture in production AD-ECUs: \textbf{Primary-Path}, \textbf{Exposure-Time-Aligned}, and \textbf{Co-Path Coexistence}. On a Dual-SoC production AD-ECU, the resulting implementation closes camera-shutter to \swc{NNPLANNER} at a mean of 296 ms within the 350 ms budget while keeping the perception-to-planner chain inside SoC 0 and removing the cross-SoC non-deterministic jitter. By letting an E2E trajectory co-run with the modular pipeline under a shared $\tau_{\text{exp}}$ reference, the architecture preserves the planner-output boundary across the eventual modular-to-E2E handover, turning the transitional period itself into a controlled migration path rather than a throwaway phase.

\section*{\refname}
\printbibliography[heading=none]

\end{document}